# Cloud-Based Autonomous Indoor Navigation: A Case Study

Uthman Baroudi[1], M. Alharbi[1], K. Alhouty[1], H. Baafeef [1], K. Alofi[1]

[1] Wireless Sensor and Robotic Networks Laboratory, Computer Engineering Department,
King Fahd University of Petroleum and Minerals, Dhahran, Saudi Arabia
{ubaroudi}@kfupm.edu.sa

**Abstract.** In this case study, we design, integrate and implement a cloud-enabled autonomous robotic navigation system. The system has the following features: map generation and robot coordination via cloud service and video streaming to allow online monitoring and control in case of emergency. The system has been tested to generate a map for a long corridor using two modes: manual and autonomous. The autonomous mode has shown more accurate map. In addition, the field experiments confirm the benefit of offloading the heavy computation to the cloud by significantly shortening the time required to build the map.

## 1 Introduction

Robots are becoming essential to our world more than ever especially when the job is critical and can endanger humans. Robots are penetrating almost every sector even our social life. Moreover, they can operate in hostile environments where human intervention is not feasible like a burning building or a radioactively contaminated site, etc. In such scenarios, robots are the perfect tool to intervene and provide help for both people inside the building and emergency crews. One of the critical issues facing emergency crews in such conditions is the internal map of the concerned building that might be not available or not up to date. Therefore, robots' capabilities can be exploited to construct a map of an unknown environment and navigate through it. This map can be used to rescue trapped people, have real-time update on changes in the surrounding environment, or locate the source of the accident.

On the other hand, producing such maps is not an easy task. It is time and resource consuming problem. However, cloud computing has opened a new venue as well as new applications. Integrating cloud computing with networked robots has been the focus of huge industrial and academic research in the recent years [11][12] [23]. Cloud computing would offer the following three fundamental benefits:
- Offloading the heavy execution of map merger algorithm and robot coordination,
- Reducing power consumption and cost requirements of the robots.
- Improving operation time and robot mobility.

Nevertheless, cloud computing may produce additional delay that may affect the mission that the robots are trying to achieve [12].

The main objective of this case study is to leverage the available robotic software tools in order to design, integrate and implement a cloud-enabled autonomous robotic navigation system. The system is assumed to be deployed in an unexplored area, so no partial maps available [22]. The robots will build a map while they are navigating the concerned area simultaneously. In addition, light video streaming is transmitted to the command center in order to facilitate online monitoring and control in case of emergency. Three different experiments have been carried out for indoor map building with video demonstration: manually operated single robot, single Robot with autonomous mapping and multi-Robot (two) autonomous mapping scenarios.

This chapter will present a detailed description of experiments design and running scenarios (i.e. with and without cloud service), and observations made during experiments. In section 2, the problem statement and system requirements are presented and discussed. Then, section 3 presents and evaluates the proposed system design including software and hardware components. In section 4, we discuss the experimental setup and obtained results. Section 5 discusses the challenges encountered during the development and experimentation. In section 6, we briefly present and discuss the related work. Finally, we conclude with lessons learned in section 7.

## 2 Problem Statement and System Requirements

### 2.1 Problem Statement

Given an unknown and unexplored area with no map available, it is required to build a 2D map for this area using multiple robots and stream video shots while the area is being explored.

### 2.2 System Requirements

- The robot must be autonomous (i.e. completely self-operated, reacting with its environment as needed).
- The system must use robots for data collection only; no processing should be done on them. This is to extend their battery lives.
- The system should use a minimum of two robots navigating and providing data simultaneously to a server on the cloud.
- The system must utilize a cloud service to do all processing needed.
- The system should produce a complete map of the environment within a building and its layout.

- The system must provide lightweight video streaming with minimum resolution of standard-definition resolution (640x480p).

## 3. System Design

In this section, we start by presenting and discussing the possible approaches to achieve the above requirements. Then, the proposed system architecture is presented. Finally, we focus on two design specific issues namely, minimum laser scan rate and minimum bandwidth requirements.

### 3.1 Possible Approaches

Table 1 shows a list of chosen possible approaches alone with their advantages and disadvantages. Selected approaches are underlined and their justifications are listed in the next section.

**Table 1: Possible Approaches**

| | Possible Approach | Advantages | Disadvantages |
|---|---|---|---|
| Number of Robots | Single Robot | ▪ Less network complexity<br>▪ Ease of management | ▪ Longer time to build a map |
| | Multiple Robots | ▪ Faster Map building<br>▪ Independent functionality | ▪ Hard to manage and coordinate navigation<br>▪ Network Complexity<br>▪ Merged maps tend to have errors |
| Robots | Quadcopter (Drone) | ▪ Mobility | ▪ Fragile |
| | Turtlebot | ▪ Availability<br>▪ Durability | ▪ Depends on external station |
| Sensors | SICK Laser Scanner | ▪ Wide field of view<br>▪ Higher accuracy | ▪ Expensive |
| | Asus Xtion Pro RGB-Depth camera | ▪ Low cost<br>▪ Availability | ▪ Lower accuracy<br>▪ Lower field of view |
| Communication | 3/4G | ▪ Wide coverage | ▪ More costly |
| | Bluetooth | ▪ Low cost | ▪ Less secure<br>▪ Low coverage area |

|  | Wi-Fi | - Good indoor coverage<br>- Reliability<br>- Availability | - Less penetration |
|---|---|---|---|
| Navigation | Manual | - Ease of implementation | - A human operator for each robot |
|  | Autonomous | - No human intervention | - Difficult to implement |
| SLAM algorithm | Gmapping | - Widely used<br>- Good accuracy | - Accuracy is not consistent |
|  | KartoSLAM | - Better accuracy | - Little documentation |
| Master Node | Robot | - Race for resources | - Reduces battery life<br>- Race for resources |
|  | Cloud | - Power-efficient<br>- Reliable<br>- Cost-effective | - Requires constant Internet connection |

### 3.2 Selected Approaches

Firstly, as per system requirements, a multi-robot approach is required and, therefore, is selected. As stated, this would help in covering a given area in a shorter period of time. Secondly, as for the robot selection, Turtlebot satisfies the stated requirements and its immediate availability made it our choice over the quadcopter option or any other option. Thirdly, as in [9], although SICK laser scanner – priced at $5,000 – is more accurate and covers a longer distance, the ASUS Xtion Pro RGB-D camera is much cheaper, priced at 150$. Nonetheless, the specifications of the laser scanner included in the ASUS Xtion Pro are sufficient for the purposes of this project. Furthermore, it is included in the Turtlebot robot package which adds the availability factor to the selection process of this laser sensor.

Fourthly, having multiple robots working with a cloud server prompts the need of a reliable communication method with medium to wide area coverage. Although 3/4G covers large areas, it is not a feasible option since it is not free and its indoor coverage is poor for exchanging heavy traffic. Bluetooth, on the other hand, is the cheapest; however, it is less secure than Wi-Fi and covers a smaller area too.

The TCP/IP network model running over a Wi-Fi network is a well-known, well-tested, and reliable model. Moreover, our workplace and testing area is well-covered with Wi-Fi access points which added the availability factor to the selection of this approach.

Fifthly, as stated in system requirements, the master node needs to be running in the cloud in order to have a power-efficient, more reliable, and more cost-effective overall solution. Also, heavy computations such as map generation and merging are now done in the cloud instead of the robots, which help in reducing power requirements for the robots and thus extend their battery life.

Finally, the KartoSLAM algorithm is selected as the simultaneous localization and mapping (SLAM) algorithm; which is used for map generation and robot localization. This is because it is more accurate in real-life experiments than other SLAM algorithms available on the Robot Operating System, including the Gmapping algorithm [8].

### 3.3 System Architecture

Having discussed some options for our system design, Figure 1 illustrates the system architecture. In the following subsections, we will discuss in details both the hardware and software components and network design.

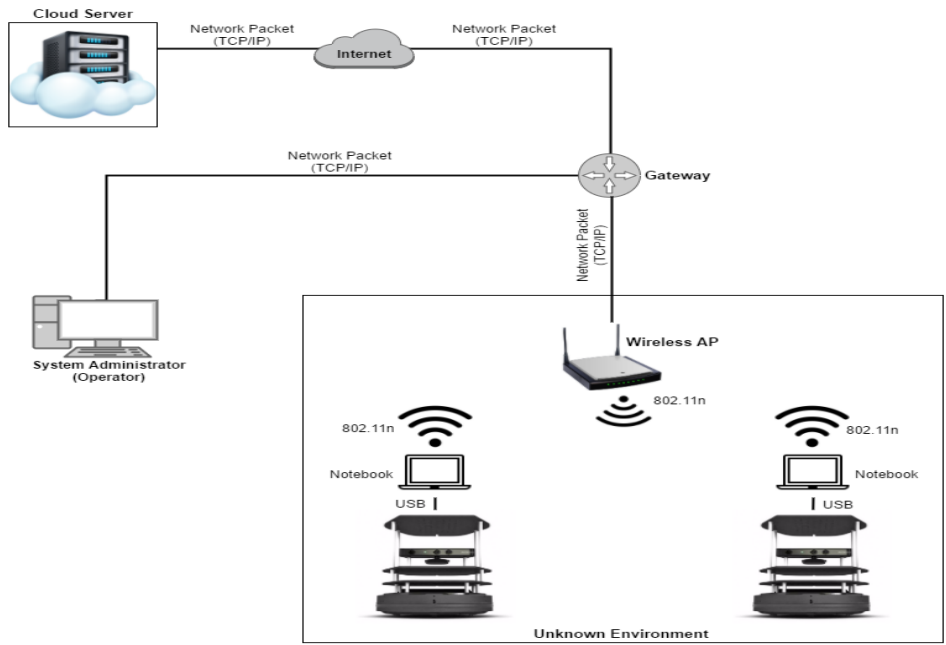

**Figure 1: System Architecture**

#### 3.3.1 Hardware Components

Figure 3 shows the complete Turtlebot setup, which has three key elements: Kobuki mobile base, ASUS Xtion Pro, and a notebook [10].

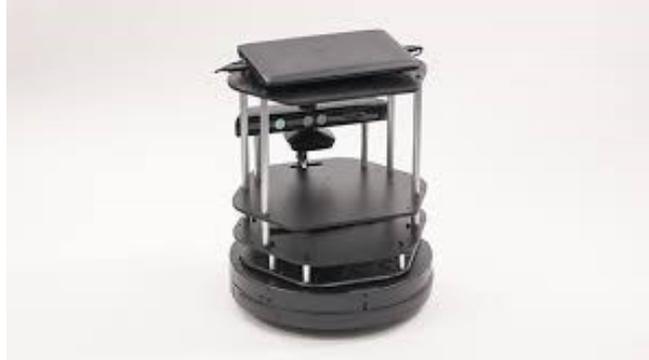

**Figure 2: : Complete Turtlebot Setup [10]**

**Kobuki Mobile Base**

This part is what makes the setup moves. The mobile base has three wheels, bump sensor, cliff sensor, LEDs, and a battery. The movement of the mobile base is controlled by the navigation algorithm. The base is connected to the notebook via USB.

**ASUS Xtion Pro**

This is a pre-installed RGB-Depth camera sensor that will provide video feed and laser information which will be used for video streaming, robot navigation, and map generation. The Asus Xtion Pro connects to the notebook via USB.

**Notebook**

The notebook has the drivers for both: the mobile base and the RGB-D camera, which enables it to control both of components. ROS is pre-installed in the notebook.

**3.3.2 Network Design**

As seen in Figure 4, the network consists of three main parts: a gateway, a wireless access point, and end systems, which includes the operator station, the cloud service, and the robots traversing the environment. Constrained by the testing environment, the system could not have an actual cloud since KFUPM's network restrict access to/from devices running within it. Instead, the cloud is simulated by a workstation working over KFUPM's local network. This workstation serves as both: a cloud service and an operator station. It is designed to run as a ROS master node in order to control the robots, coordinate between them, provide services to them, log their activities, collect data from them, construct a map of their environment and view it, update an existing map, and save maps if needed.

This design approach necessitates discussing two important issues namely, why a single cloud master versus multi-master design and the effect of using local workstation instead of a real cloud service. For the first issue, having a cloud master will consolidate the process of coordination among different robots and help in better and faster response [12][19]. On the other hand, having a single cloud master constitutes a single point of failure, but it is rare to happen.

In regard to the later issue, this design decision will not affect the system in any way except that it reduces latency and hence, makes the overall system perform slightly faster. All other functionalities are not altered should the system use an actual cloud service. Table 2 shows the average latency assuming Azure cloud service is used. This level latency has minimal effect on our concerned application here in this case study. However, if the application is delay sensitive such as self-driving, the designer should be very careful in adopting such approach.

**Table 2: Average latency between KFUPM and closest Azure data centers [21].**

| UK West ( Cardiff ) | 115 ms |
| North Europe ( Ireland ) | 126 ms |
| West Europe ( Netherlands ) | 136 ms |

**3.3.3 Minimum Laser Scan Rate**

This factor is important as it plays two roles: ensuring quality of transmitted scenes and possible congestion on the used network. Therefore, we need to strike a balance between these two roles.  The laser rangefinder's is capable of capturing 30 scans/second, the system, however, does not require all of them to build a proper map of the scanned area. Hence, we reduce the laser scan rate and at the same time relieve the network from excess traffic.

In order to determine the minimum laser scan rate, a maximum time between scans ($\Delta t$) must be estimated. We denote $\Delta t$ as inter-scanning period. Estimating this parameter depends on a number of factors such as the robot speed, computing power, map construction software, etc.  For this case study, these parameters are fixed as determined above by the chosen hardware/software specifications. Hence, we are just going to estimate the minimum rate through experimentation. It was found that the average maximum distance between scanned objects and the camera such that the mapper software is able to build an accurate map is about $D = 1.5$ meters. In addition, it is found that the maximum speed of the robot is $V = 0.65$ m/s.

Thus, when the robot is moving at maximum speed, a frame every:
$$\max\{\Delta t\} = \max\{D\}/V \tag{1}$$
So, the maximum inter-scanning time ($\Delta t$) is 2.308. Therefore, the minimum needed scan rate is $1/2.308 = 0.44$ scan/sec.

Although this is the minimum required scanning rate, it is recommended by the map building package to have a minimum of 1 scan/second. Furthermore, in order for the robot to be able to avoid obstacles properly, a laser scan rate of 10 scans/second is recommended for robots travelling at the speed of 1.5 m/s or less [4]. Hence, in this case study, we set the laser scanning rate (denoted as *mni_SR*) to 10 scans per second.

### 3.3.4 Bandwidth Requirements

Network bandwidth is a scarce resource and it should be used efficiently. Since there is going to be a cloud service in communication with robots in the field, the bandwidth utilization must be minimized as much as possible. The following calculations provide an estimation of the required network bandwidth.

The network is going to be used for two purposes: video streaming, and control commands. Since control commands are typically small packets, so it will be ignored in our computation. Therefore, we will consider only the requirements of video streaming.

Each robot has two cameras: an RGB color camera for video streaming and a depth camera that is used to build the maps. Both cameras have a resolution of 640 X 480 pixels per frame; denoted as $F$. The RGB camera, however, needs 3 bytes for each pixel (denoted as $L_{RGB}$) while the depth sensor needs 1 byte for each pixel (denoted as $L_D$). Both sensors stream 30 frames per second, however, as stated previously, the laser scanner is set to 10 frames per second. Therefore, the estimated bandwidth ($B$) is:

$$B = F[30(L_{RGB}) + 10(L_D)] \qquad (2)$$

$$\begin{aligned} B &= 640 \ X \ 480 \ pixels/frame \ X \ [(30 \ frames/second \ X \ 3 \ bytes/pixel) \\ &\quad + (10 \ frames/second \ X \ 1 \ byte/pixel)] \\ &= 30{,}720{,}000 \ bytes/second \\ &= 29.30 \ MB/s \ per \ robot \end{aligned}$$

As mentioned earlier, network bandwidth must be minimized as much as possible. Therefore, a compression technique is needed to reduce the bandwidth. JPEG compression with a ratio of 23:1 would result in a significant bandwidth improvement: (29.30 MB/s)/23 = 1.28 MB/s per robot. As a result, the system will need a minimum bandwidth of 1.28 MB/s per robot.

### 3.3.5   Software Architecture

This section discusses the fundamental structures of software and the messages exchanged between different software components. Figure 3 illustrates the software structure diagram. The main software component in our system is the Robot Operating system (ROS); all other software components are built on top of it. In addition, an ROS package called "nav2d" [2] has been used. It is a 2D-navigation package that meets system requirements of autonomous robot navigation and mapping. Furthermore, the system has three extra software components: the camera driver, the mobile-base drive, and the stream view. In the following subsections, we will discuss each component in detail.

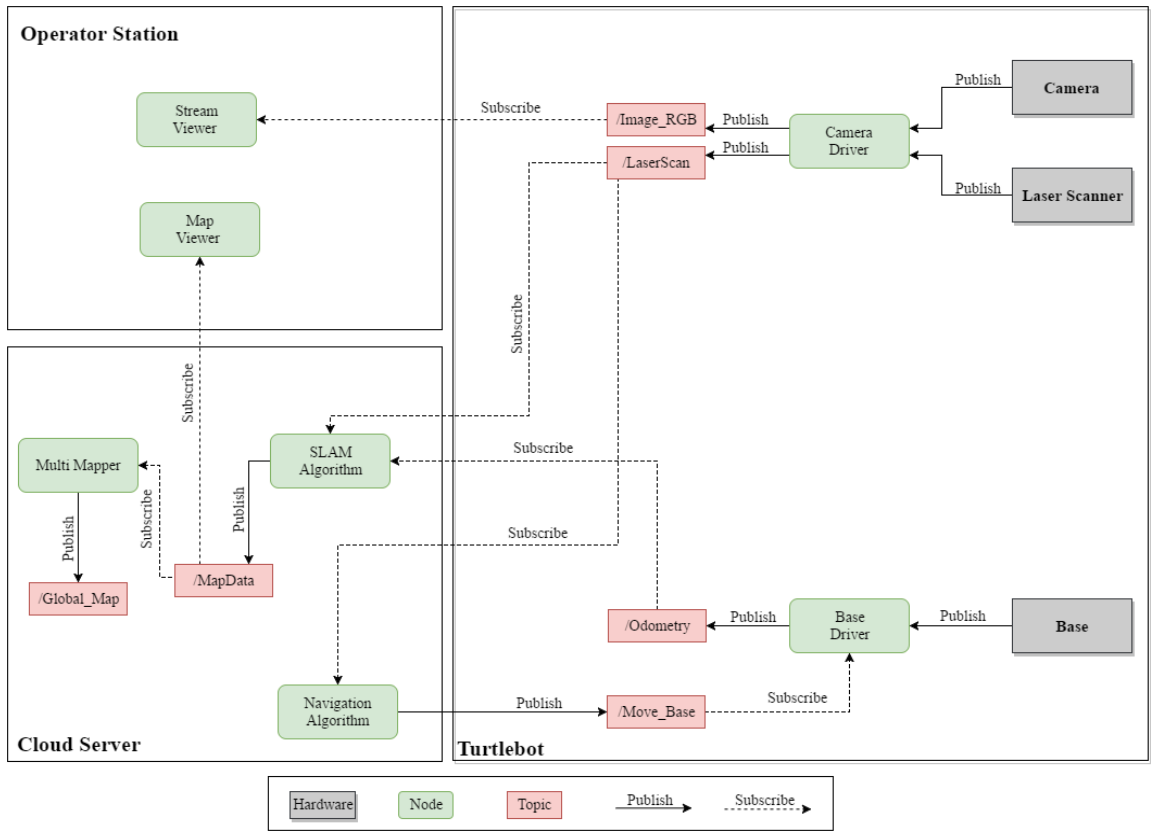

**Figure 3: Software Structure Diagram**

**Robot Operating System (ROS) [1]**

ROS is an open-source, distributed operating system, which provides services such as hardware abstraction, low-level device control, implementation of common functionalities, message passing between processes, and package management. The main goal of ROS is to create a general framework that eases the implementation of robot applications for different types of robots. Different robots have varying hardware properties, which make implementation and code reuse more difficult and less frequent. ROS aims to solve this problem by introducing sets of libraries that provide certain functionalities that are common in a certain type of robots.

ROS uses a publish-subscribe messaging-scheme as a communication model between nodes. However, a modern OS cannot ignore the need for a request/reply communication model; this model is used between services. In ROS, when multiple robots are working cooperatively, all robots should be connected to a *master node* or a ROS *master*. This master node is responsible for communication coordination between other nodes and therefore, provides services such as name registration and lookup information. Figure 4 summarizes the *four* ROS fundamental concepts [5].

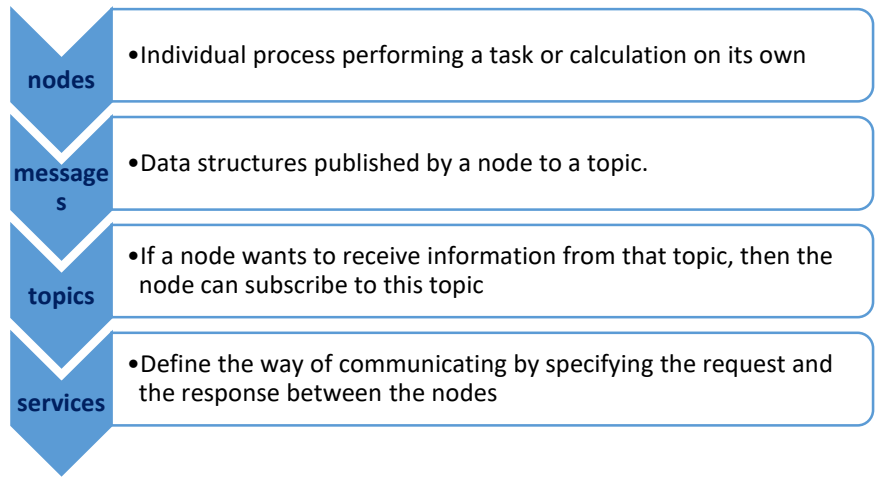

**Figure 4: A summary of ROS four fundamental concepts.**

**Nav2d Package**

This package has five main nodes as depicted in Figure 5: mapper, explorer, navigator, operator and localizer.

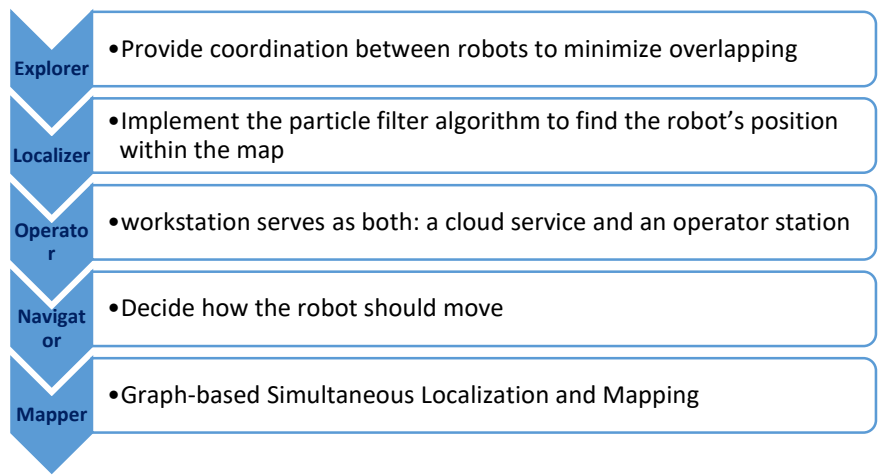

**Figure 5: A summary of nav2d package main nodes [2].**

**Explorer:**

This package provides coordination between multiple robots in order to minimize overlapping. An unexplored area of a map is called *frontier*. This node has three different exploration strategies for these frontiers:

- **Nearest frontier**: performs simple nearest-frontier exploration without any coordination between the robots.
- **Multi-robot nearest frontier**: extends the nearest-frontier strategy to a multi-robot navigation system by providing coordination between robots using a synchronized wave-fronts at each robot's position.
- **MinPos**: a novel frontier allocation algorithm for multi-robot exploration.

As the number of robots increases, the exploration time for both algorithms decrease. Nonetheless, the MinPos is more time-efficient than the nearest-frontier at all scenarios [6]. As a result, it was chosen as the coordination mechanism between the robots over the nearest-frontier algorithm. This node and the navigator node are verified together through simulation, see Figure 6.

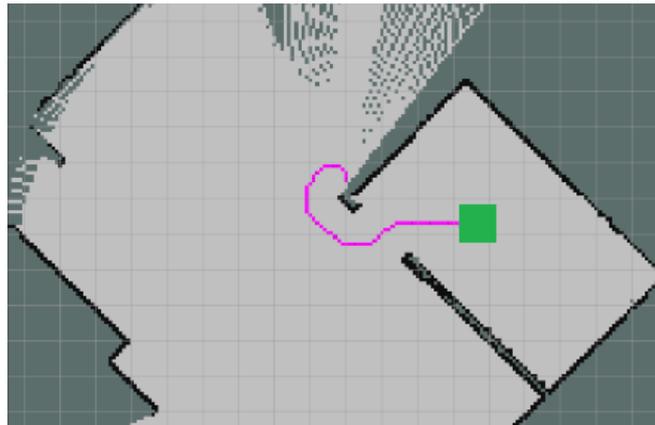

Figure 6: The green square represents the robot. The exploration method is closet-frontier which is passed from the explorer node to the navigator. The navigator plans the path for the robot which is shown by the purple line.

**Localizer**

The localizer node implements the particle filter algorithm to find the robot's position within the map [7]. This node is only used when the robot wants to navigate a fully pre-built map, or cooperate in constructing a partially-built one.

**Operator**

The operator node is a lightweight, purely reactive obstacle-avoidance module for mobile robots moving in a planar environment. It takes motion commands from the explorer node, evaluates them based on a predefined set of motion primitives (e.g. the size of the robot), and outputs control commands to the robot hardware avoiding obstacles in front of the robot. Figure 7 demonstrates a simulation tutorial experiment that we run to verify the proper functionality of this software module by avoiding obstacles.

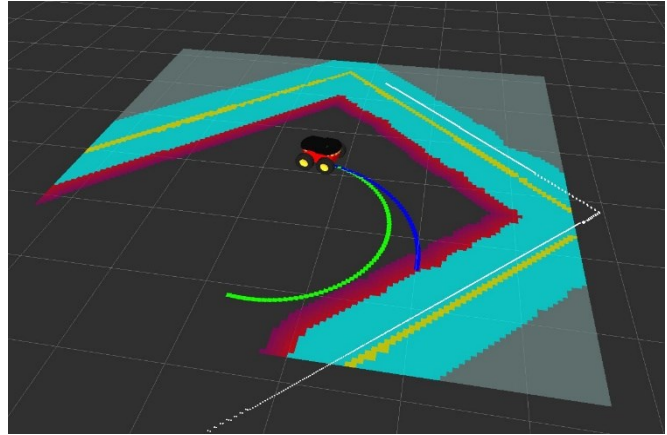

**Figure 7: Blue and red points are obstacles. The dark-blue line emitted from the robot is the desired path, while the green line is the corrected path.**

**Navigator**

The navigator node decides how the robot should move – if it were to move autonomously – within its environment. However, it requires the current map, and the position of the robot within this map to generate a navigation plan. This node cooperates with the Operator node and the Explorer node in order to achieve autonomous navigation. This node provides four services to other nodes:

- MoveToPosition2D: requires a set of coordinates and moves the robot to these given coordinates
- StartMapping: requires no parameters and starts building the map
- StartExploration: requires no parameters and starts exploring the environment in accordance with the exploration method chosen (see explorer node)
- Localize: requires no parameters and starts localizing the robot within its map.

**Mapper**

The mapper uses a Graph-based Simultaneous Localization and Mapping (Graph-SLAM or GSLAM) algorithm; a modified version of the SLAM category. The function of the GSLAM is to construct or update a map of an unknown environment and keeping track of the location of the robot within the produced map; this is essential since no external referencing techniques such as GPS are available. GSLAM uses different types of sensors to produce visual features of the surrounding environment. A laser-based GSLAM algorithm is used to produce a 2-D grid map from laser input and odometry information provided by the camera driver and the mobile-base driver, respectively.

This node uses the Open Karto mapping library developed by SRI International. Unfortunately, there is little documentation about the Open Karto library [8]. The mapper also supports map-building using multiple robots. This is possible by keeping track of the positions of all robots and subcriping to the laser feeds provided by them. This node and the localizer node are verified together through simulation as seen in Figure 8-13.

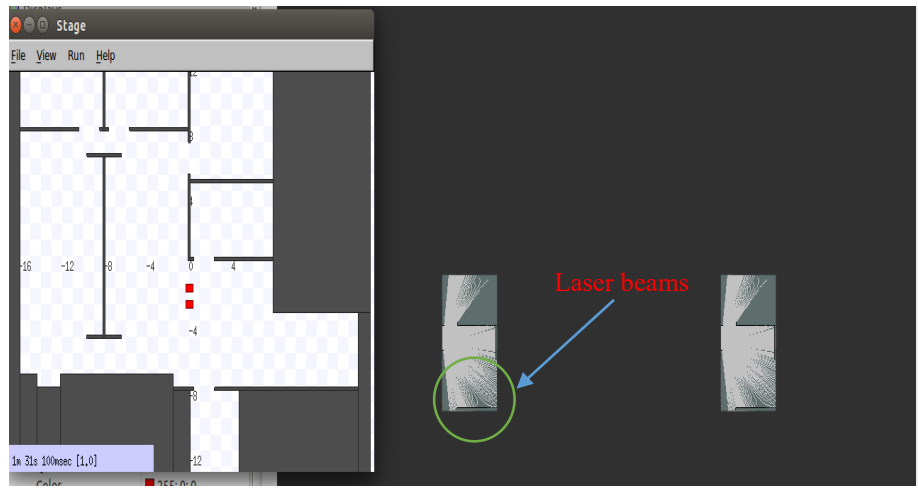

**Figure 8:** Robot 1 starts its mapper; the two red dots indicate the robots initial positions. One robot starts anchor itself to the map using laser beams (i.e. out rays).

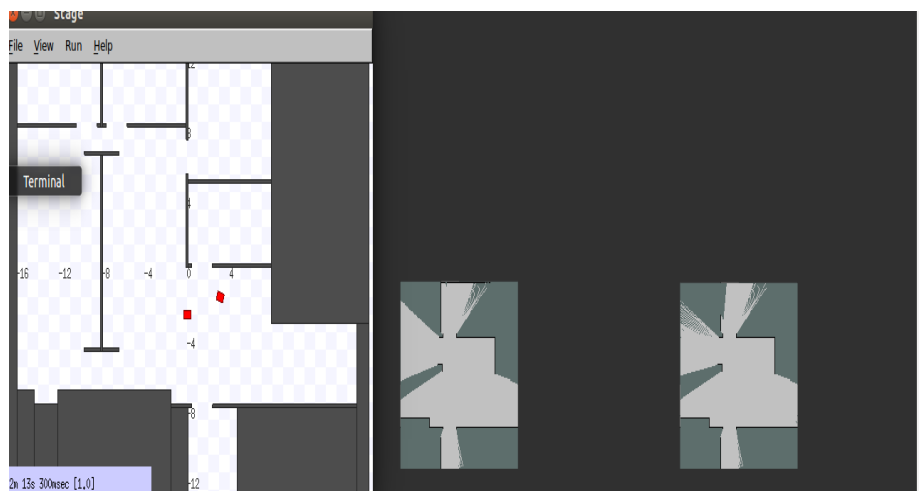

**Figure 9:** Robot 1 mapper is expanding the map; we can observe that Robot 2 still waiting in its initial position.

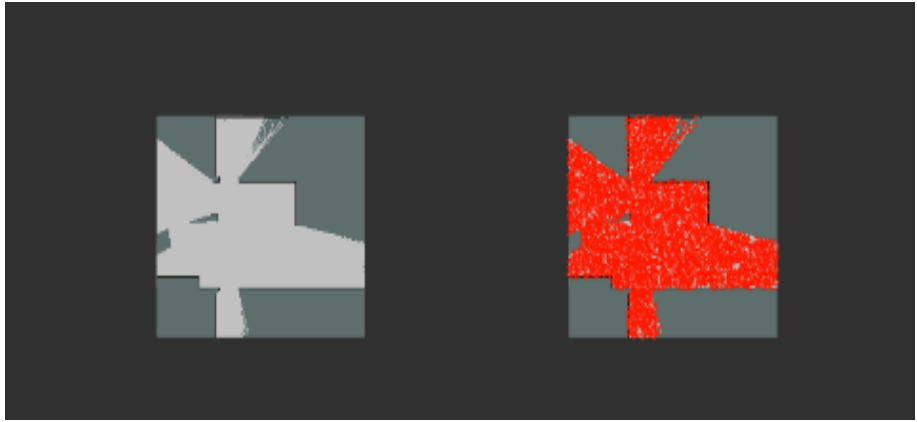

**Figure 10:** Robot1 shares its map with Robot 2 (right map). Then, Robot 2 initializes a particle filter to localize itself within the map; the right red dots are the random particles generated by the filter.

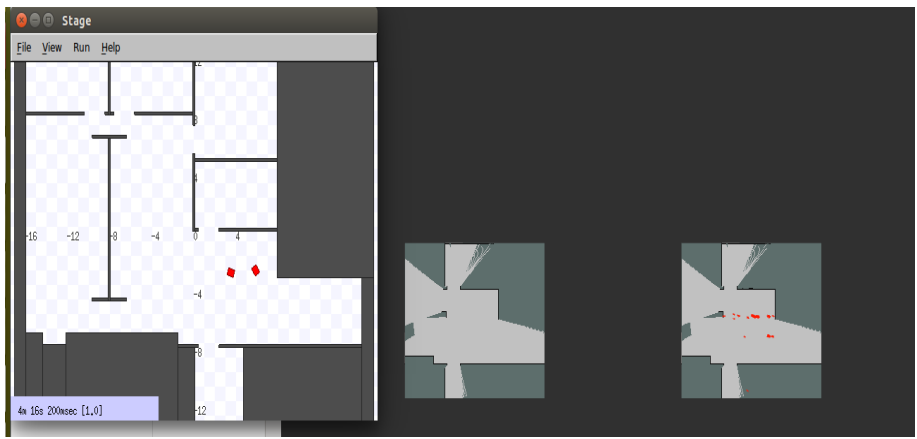

**Figure 11:** Particles (red dots) start to converge

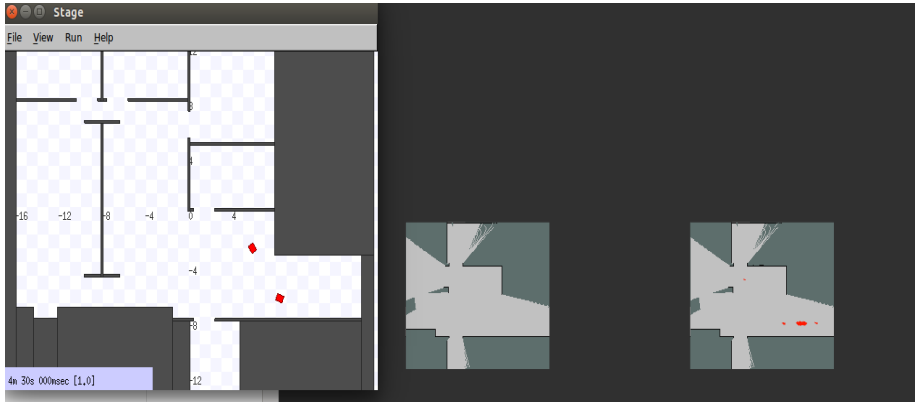

**Figure 12:** Particles start to converge into one place

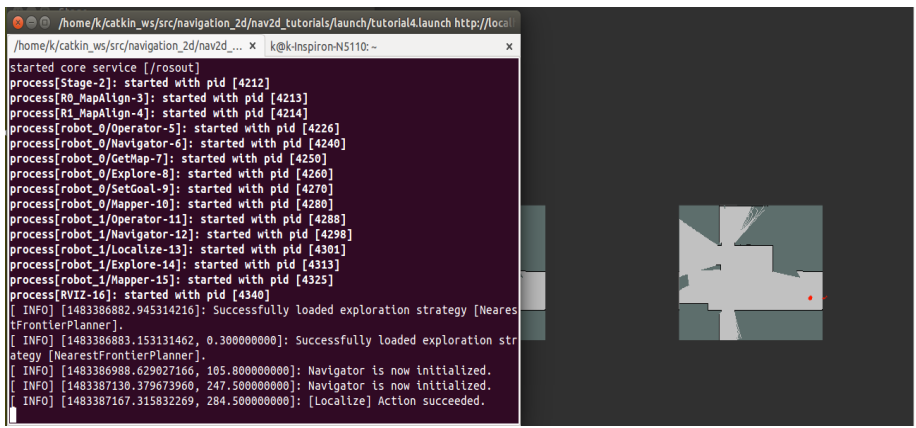

**Figure 13:** Localization of robot 2 succeeded and Robot 2 localizes itself in the map (right map) and starts moving to explore the area and expand the map collaboratively with Robot 1. These steps are displayed on the left box of ongoing actions.

# 4   System Integration and Testing

### 4.1 Experimental Setup

After testing each component separately, both hardware and software components, and verifying that each component is working properly and as expected, we integrate all system components together as illustrated in Figure 14. We have begun by setting the workstation so that it controls the robots and process the acquired data from them.

The version of Robot Operating System (ROS) included with each Turtlebot is called "Indigo." It requires a special version of Linux distributions in order to operate properly. Upgrading to newer versions is not feasible since the needed packages are not available on them. Therefore, the only software specification is the use of Ubuntu 14.04 long-term support Linux distribution.

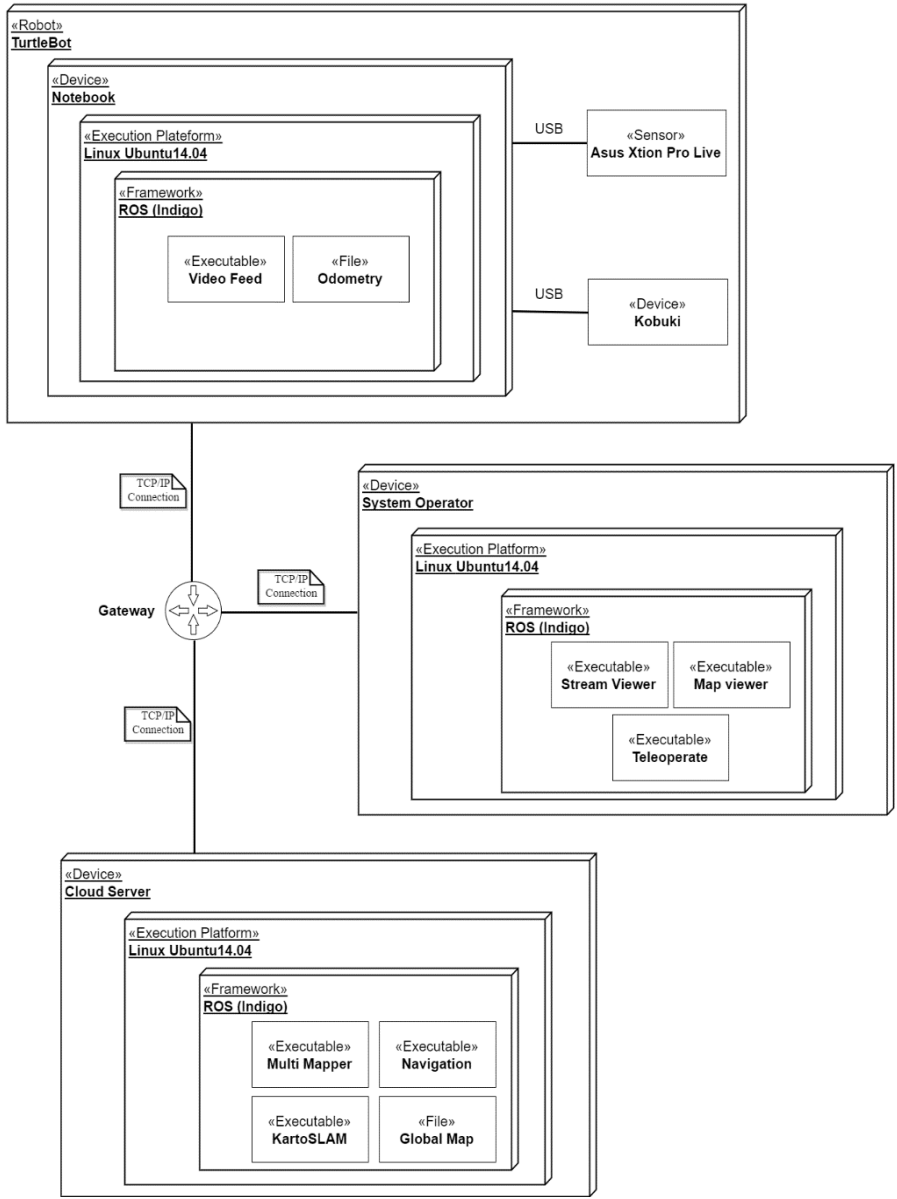

**Figure 14: System components integration**

In the following, we will demonstrate our navigation experiments for manual, autonomous with a single robot and with two robots. For single robot scenario, the experiment started by placing the robot at the entrance of the corridor as clearly shown in the video. For two-robot scenario, one robot is placed at the entrance while the second robot is placed in the middle; this will reduce the total travelled distance and minimize overlapping images.

At the point of time, Robot#1 has started the exploration and mapping process. Then, Robot#2, using the shared image and particle filter, will be able to localize itself and extend the map.

**4.2 Performance Evaluation**

**4.2.1  Manual exploration using one robot**

This test has been performed so that it involves full human interaction for the exploration process. It can be noticed that although it is faster than the autonomous scenario with a single robot, the obtained map has many incorrect edges.

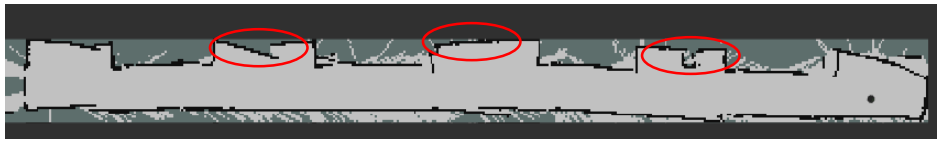

**Figure 15:** The Corridor map generated manually; red circles point to incorrect edges.

**4.2.2  Autonomous exploration using one robot**

This test has been done using nearest frontier exploration method autonomously with a single robot. The obtained dimensions, corners, edges are very close to the real map. The experiment can be viewed on this link: https://youtu.be/rk8hCgbV6AM. As illustrated by this video, the system is able to capture very detailed stream of images which can be used by the operator for rescue operation, for instance.

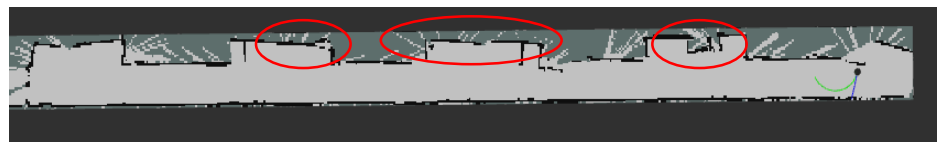

**Figure 16:** The Corridor map generated autonomously; better captured edges

### 4.2.3 Autonomous exploration using two robots

This test compromises of using two robots to generate a map. It is the most accurate map which obtained in the shortest operation time (37% drop in time compared to manual single robot scenario and 52% drop compared to autonomous single robot scenario) at the expense of more data transmission. This is attributed to the redundant images sent by both robots. The experiment can be viewed on this link: https://youtu.be/HeBHDXDElwQ.

**Table 3:** Performance Comparison for the three exploration Scenarios; the single robot scenario is the baseline for the improvement comparison **(+)** means enhancement while **(-)** means degradation.

| Test Environment: a long corridor | Single robot scenario | | | Two robots scenario | |
|---|---|---|---|---|---|
| | Manual | Autonomous | | | |
| | | | % | | % |
| Completion Time (Sec) | 430 | 570 | -33 | 270 | +37 |
| Travelled Distance (m) | 50.3 | 55 | -9.3 | 55 | -9.3 |
| Data Transferred (MB) | 224 | 290 | -29.4 | 450 | -100 |
| Avg. speed (m/s) | 0.5 | 0.4 | n/a | 0.4 | n/a |

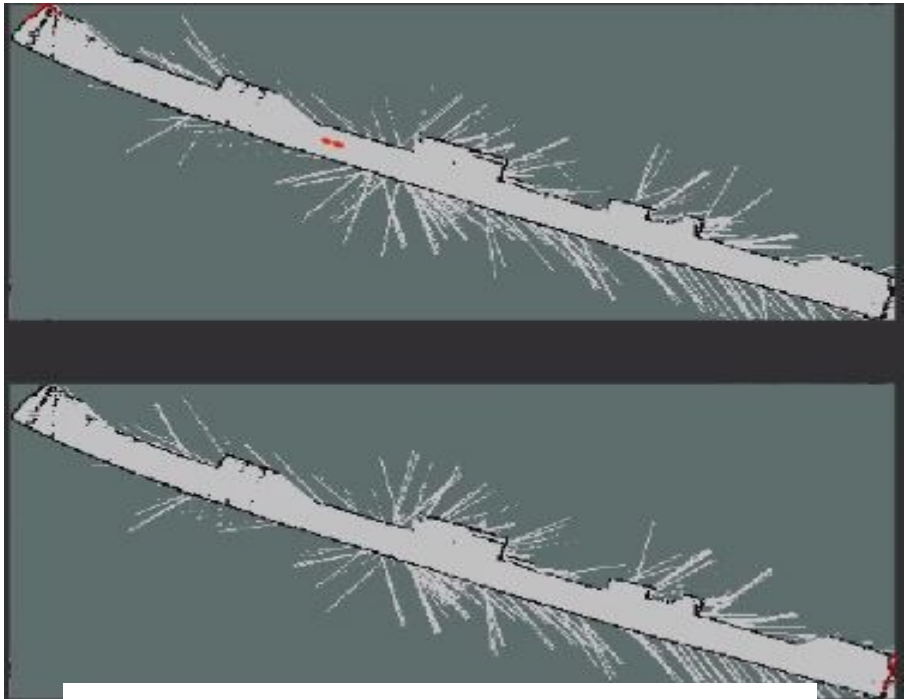

**Figure 17: The Map of a corridor generated by two robots**

# 5 Technical Challenges

During the development of this work, we have experienced several challenges and problems. We will discuss the major and most important problems.

## 5.1 ROS

We have faced many problems such as lack of good documentations if any and a small community of developers. This has made the journey even harder for us. We tried to reach out to some experts in ROS, but little responses were received.

Moreover, a huge and important library in ROS is one called "tf," or the transform library. This library is responsible for keeping track of multiple coordinate frames in the system over a specified period of time and transforming between them. Each directional part of ROS has a coordinate frame; being multiple moving robots base each with their coordinate frame which consists of, for instance, its X, Y, and Z coordinates alongside some special information such as odometry, or a robot arm that has also X, Y, and Z coordinates accompanied with an angle. This library is complicated and huge, but it was born out of need. In the early days of ROS, the transformation process was manual and each developer had to implement it on their own; keeping track of each individual frame in their system, and there are usually tens of them sometimes exceeding a hundred.

Moreover, running multiple robots means that each robot had to be operating in a separate namespace and tf prefix. Namespaces are used to publish topics, nodes, and services while tf prefixes are used to publish coordinate frames for each individual robot. This is not an easy task to accomplish since almost all ROS packages used in our system made the assumption of running always in a global namespace, where no more than one robot is operating. This, of course, introduces a conflict once multiple robots are deployed. Because each robot would publish conflicting information over the same namespace and tf prefix. Therefore, we need to fix this problem and make each robot allocating its own unique namespace and tf prefix and not use the global one.

## 5.2 Physical constrains

The ASUS Xtion Pro converts depth images to laser scan and it has a 3-meter range, which is not appropriate for mapping and navigation application. Because of this short range, the robot has to travel longer distance and change its direction frequently causing a shift in the odometry and this error propagates to the whole map. In addition, the placement of the RGB-D sensor on the base is very crucial since any small shift in it would add a further shift to the odometry in the map.

## 6 Related Work

The advancement in sensing, communications technologies and software engineering paves the way for unprecedented applications. Figure 18 depicts the time line for networked robotic applications. It is obvious from Fig. 18-the slow development of robotic application until beginning of 2010 when a huge advancement has been introduced exploiting the cloud services. In addition, the introduction of industry 4.0 gave this branch of knowledge another push which introduces another dimension that is huge collected of data by IoT devices and how they can contribute to the robustness of the mission carried by robots via cloud computing.

One of these applications is the networked robots and internet of robotic things that can perform multi-task duties simultaneously using computing facilities which are spread over the world [11][12][14][15][17][19][23][27]. Simultaneous localization and mapping (SLAM) algorithm is old problem that has been the focus of huge research efforts since 1990. SLAM is the core of many advanced real-time applications, such as self-driving cars, rescue operations, surveillance, etc.

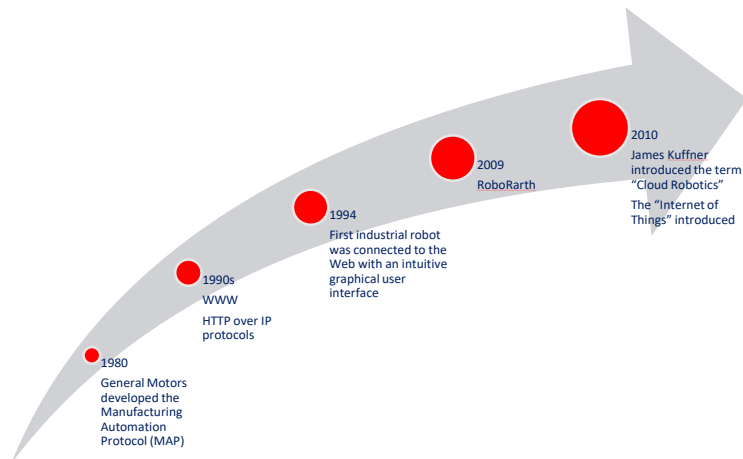

(a)

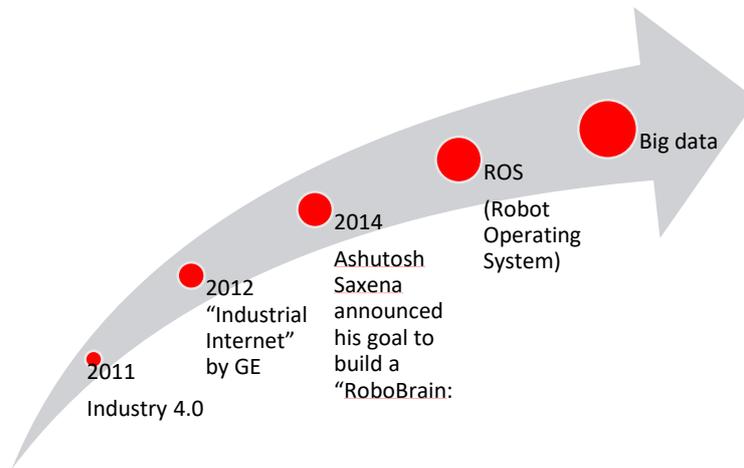

(b)

**Figure 18: Timeline of robotic applications development extracted from [23].**

The authors in [24] have identified six attributes for SLAM algorithms to be suitable for self-driving applications. These attributes are: Accuracy, Scalability, Availability, Recovery, Updatability, Dynamicity. However, SLAM requires tremendous computing resources that local computing facility is not an option and not to mention the lack of other attributes. Therefore, in order for new real-time applications to be materialized and fulfill their missions, cloud enabled SLAM is the foreseen solution. Nonetheless, this solution still needs a long journey to mature and satisfy the above mentioned attributes.

Considering the cloud software infrastructure, we can recognize three levels. Infrastructure as a Service (IaaS) is the lowest level, where operating systems are provided on machines in the Cloud. Platform as a Service (PaaS) such as ROS constitutes the second level, where application frameworks and database access are provided, but with restricted choices of programming languages, system architectures, and database models. The third level is Software as a Service (SaaS) [23].

Furthermore, open source software is witnessing a good acceptance in the robotics and automation community. For example, ROS, the Robot Operating System has been ported to Android devices [16][23].

In this chapter, we design a cloud enabled system using ROS as our software platform where a master ROS is hosted in the cloud. We opted for this option for two main reasons: better coordination among the deployed robots using a single controller and offloading the computing complexity of running SLAM to the cloud where Abundance of resources is available with affordable cost. This option is very suitable for our application (building a map for known area).

On the other hand, having a single master ROS node may suffer the following problem considering ROS as PaaS. First, when a single ROS master node is employed, services and nodes may be in conflict with having the same name which requires a careful design of namespace for ROS nodes, services and topics. This issue becomes worse with the large number of robots. The lack of scalability is another design issue when the single ROS master is responsible for coordinating among multiple robots simultaneously [25]. ROSLink is suggested to be used in this case [25].

## 7   Conclusion

Autonomous exploration and collaborative map building on a remote computing resource by using raw data is a very useful feature of multi-robot system. In this chapter, we have successfully designed, integrated and implemented a cloud-based autonomous navigation using mutli-robot system. The system used multiple robots to traverse a concerned area and build very accurate map. The system was tested thoroughly for different scenarios namely, manual navigation and autonomous navigation. Autonomous navigation with two robots have shown faster and more accurate map compared to manual navigation. Furthermore, offloading the computation task to a workstation in a cloud had saved the system both extensive local energy consumption at the robot side besides quick update for generated map.

It is observed that the amount of exchanged data between robots and the cloud is almost doubled for the case of two robots compared to a single robot case. For future work, new approaches need to be developed to minimize the amount of exchanged data and test the developed system under different environments. In addition, we will investigate the impact of cloud response time using a larger number of robots.

# Acknowledgment

The authors would like to acknowledge the support provided by the National Plan for Science, Technology and Innovation (MAARIFAH) - King Abdulaziz City for Science and Technology through the Science & Technology Unit at King Fahd University of Petroleum & Minerals (KFUPM), the Kingdom of Saudi Arabia, award project No. 11-ELE2147-4.